\PassOptionsToPackage{numbers,sort&compress}{natbib}
\documentclass[aps,onecolumn,11pt,notitlepage,superscriptaddress,nofootinbib]{revtex4-2}

\usepackage[utf8]{inputenc}
\usepackage[T1]{fontenc}
\usepackage{url}
\usepackage{booktabs}
\usepackage{amsfonts}
\usepackage{nicefrac}
\usepackage{microtype}
\usepackage{xcolor}
\usepackage{graphicx}
\usepackage{amsmath}
\usepackage{mathtools}
\usepackage{bm}
\usepackage{placeins}
\usepackage[hidelinks]{hyperref}
\usepackage[capitalise]{cleveref}

\AtBeginDocument{\linespread{1.213}\selectfont}

\makeatletter
\renewcommand{\p@subsection}{}
\renewcommand{\p@subsubsection}{}
\makeatother
\AddToHook{cmd/appendix/after}{%
}

\makeatletter
\renewcommand{\auto@bib@empty}{\global\let\auto@bib\@empty}
\makeatother

\begin{document}

\title{Power-Law Correlations Without Apparent Temperature-Tuned Criticality in Fixed-Tree Hierarchical Languages}

\author{Marcel K\"uhn}
\email{mkuehn@itp.uni-leipzig.de}
\affiliation{Institute for Theoretical Physics, Leipzig University}
\affiliation{Center for Scalable Data Analytics and Artificial Intelligence (ScaDS.AI), Leipzig University}

\author{Bernd Rosenow}
\affiliation{Institute for Theoretical Physics, Leipzig University}
\affiliation{Center for Scalable Data Analytics and Artificial Intelligence (ScaDS.AI), Leipzig University}

\begin{abstract}
Sampling temperature can substantially reshape autoregressive generation, but it is unclear when temperature changes in long-range correlations reflect a genuine phase transition rather than a smooth change in statistics. We study this question in a controlled analogue of temperature sampling: a finite-depth probabilistic context-free grammar on a fixed hierarchical tree, where exact next-token conditionals are computed by belief propagation and then rescaled by a softmax temperature before autoregressive sampling. We measure two-point mutual information between tokens as a function of separation. Across the tested temperatures and grammar ensembles, mutual information remains well described by a power-law envelope. Temperature changes the correlation magnitude and fitted exponent, but within the accessible system sizes we find no long-distance plateau characteristic of an ordered phase and no clear crossover to short-range exponential decay.
An approximate mapping to an Ising model on a tree explains how exponential decay in tree distance becomes a power law in token distance, allowing power laws to persist without tuning to a special decoding temperature. This controlled baseline clarifies which temperature effects can arise from hierarchical structure alone, before adding training, approximation error, or neural architecture.
\end{abstract}

\maketitle

\section{Introduction}
\label{sec:introduction}

Natural language has structure on many length scales. Nearby words are constrained by local syntax, but sentences, paragraphs, and documents also exhibit topical and semantic dependencies over much longer ranges. One concrete indicator of this multiscale organization is the presence of long-range correlations in natural language~\cite{ebeling1994entropy,altmann2012origin,mikhaylovskiy2023autocorrelations}. This matters directly for language modeling: a model that captures only local patterns misses an essential part of long-context structure~\cite{lin2017critical}.

A standard observable of long-range pairwise dependence is the two-point mutual information, which measures how much information one word carries about another word at a given separation. Empirically, natural language exhibits an approximate power-law decay of such correlations with distance~\cite{ebeling1994entropy}. This is distinct from Zipf's law, which is a one-point statistic of word frequencies rather than a two-point statistic of dependencies across positions~\cite{piantadosi2014zipfs}. A common explanation for power-law correlations in language is hierarchical organization~\cite{lin2017critical,altmann2012origin,sainburg2019parallels}. This idea has motivated recent studies that use hierarchical models, including probabilistic context-free grammars (PCFGs), as controlled testbeds for understanding learning in neural networks. Some of this work connects long-range correlations to neural scaling laws~\cite{hestness2017deep,kaplan2020scaling,hoffmann2022an,cagnetta2024correlations,cagnetta2025zipf}. Related work further shows that neural architectures can learn inference or parsing procedures in hierarchical data, including image models~\cite{malach2018provably,mei2025unets} and transformers trained on hierarchical languages or formal grammars~\cite{garnier2024hierarchy,zhao2023transformers,allen2025cfg}.

Power-law correlations are also familiar in statistical physics, where they often appear near critical points that separate qualitatively distinct phases. In a thermodynamic phase transition, a control parameter such as the Boltzmann temperature changes the macroscopic behavior of a system. For example, a ferromagnet can pass from an ordered phase with aligned spins to a disordered phase with vanishing magnetization. A critical point is not merely a threshold; it is defined by singular behavior in the infinite-system limit. Similar language has increasingly been used in machine learning, where increasing scale can lead to abrupt qualitative changes in behavior~\cite{wei2022emergent}. Examples include grokking, which has been described and analyzed in phase-transition terms~\cite{power2022grokking,zunkovic2024grokking}, and diffusion models, where denoising increasingly corrupted images can exhibit genuine transition phenomena~\cite{sclocchi2025diffusion}.

In autoregressive text generation, a natural control parameter is the sampling temperature
which rescales the logits in the softmax output and thereby, like the Boltzmann temperature of an equilibrium distribution, controls how sharply each next-token distribution is peaked.
It has been observed that low temperatures can produce repetitive outputs, high temperatures can produce incoherent outputs, and intermediate temperatures predominantly yield natural text statistics, including long-range structure~\cite{lippi2019lstm}. Together with the association between criticality and power-law correlations, this leads to a natural question: do autoregressive language models generate power-law-correlated text because they operate near a critical decoding temperature? Empirical studies have begun to examine temperature effects in actual language models, both through output statistics~\cite{arnold2024phase,mikhaylovskiy2025zipf,kempton2025distortion} and through downstream performance or reasoning behavior~\cite{renze2024temperature,xu2022systematic,zhu2024hot,pursnani2023performance}. 
Some works report numerical evidence for genuine critical behavior in generated text under temperature shifts~\cite{bahamondes2023phase,nakaishi2024critical}. However, the question remains debated. In trained neural language models, the observed statistics combine the data distribution, approximation error, architectural biases, optimization effects, and the sampling procedure. What is missing is a controlled baseline in which the underlying language distribution is explicit, the exact next-token conditionals are known, the temperature can be varied cleanly at decoding time, and training effects are absent. Such a baseline cannot replace experiments on trained language models, but it can identify which phenomena already follow from hierarchical structure and temperature-rescaled autoregressive sampling.

\paragraph{Our Contributions}
\begin{itemize}
    \item We study temperature-rescaled autoregressive sampling in a hierarchical language model: a tractable finite-depth probabilistic context-free grammar (PCFG) on a fixed tree that provides a controlled baseline with exact next-token inference.
    \item We show numerically, using sampled sequences, that two-point mutual information remains well described by a power-law decay across a broad temperature range in this model family.
    \item We show that this behavior is robust across multiple rule ensembles of the grammar, while the detailed temperature response depends on the support structure and heterogeneity of the rules.
    \item We give a tree-based statistical-mechanics interpretation, via an approximate mapping to an Ising model on a tree, for why finite-temperature correlation plateaus and short-range exponential decay do not emerge in this setting.
\end{itemize}

The rest of the paper is organized as follows. In \cref{sec:setup} we define the hierarchical language model and our analysis setup. In \cref{sec:related_work}, we discuss empirical work on temperature effects and related hierarchical language models. In \cref{sec:numerics} we present the numerical temperature analysis. In \cref{sec:ising_mapping} we interpret the persistence of power-law correlations using the geometry of trees and an approximate Ising model analogy. Finally, \cref{sec:discussion} discusses implications, limitations, and directions for extending the controlled baseline to richer observables and model classes.

\section{Hierarchical model and numerical protocol}
\label{sec:setup}

\begin{figure}
  \centering
  \includegraphics[width=0.31\textwidth]{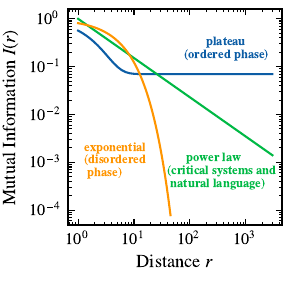}
  \hfill
  \includegraphics[height=0.39\textwidth]{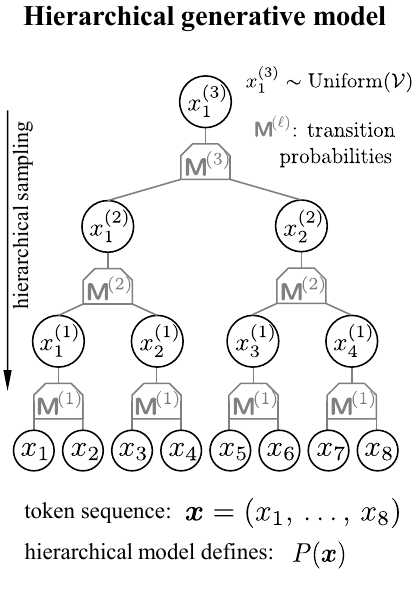}
  \hfill
  \includegraphics[height=0.39\textwidth]{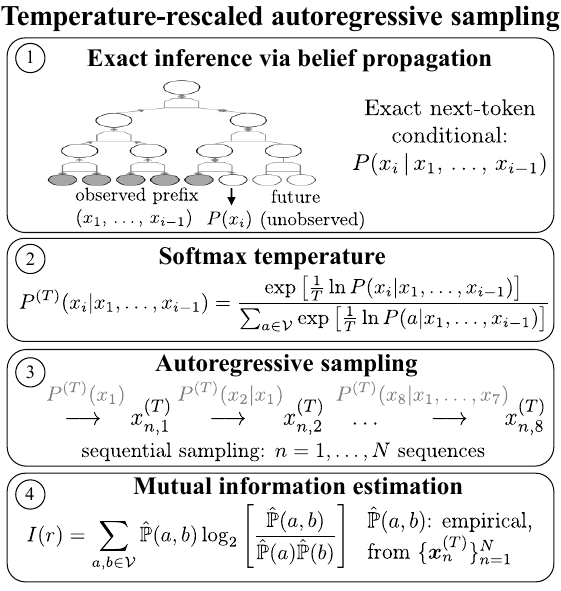}
 \caption{Left: Schematic plot of different mutual-information behaviors. In the 2D Ising model, an ordered phase would show a plateau in the mutual information, whereas a disordered phase would show exponential decay. At criticality, correlations decay as a power law. Natural language likewise exhibits an approximate power-law decay of correlations. We ask whether such decay requires tuning to a special sampling temperature. Center: Structure of the synthetic language for $s=2$ and $L=3$. Right: Protocol for autoregressively sampling text from this language at a modified temperature.}
\label{fig:concept}
\end{figure}

\paragraph{Sequences and mutual information.}
For our controlled setup, we consider discrete sequences $\bm{x}=(x_1,\ldots,x_d)$, with $x_i\in\mathcal{V}$, of fixed length $d$ over a vocabulary $\mathcal{V}=\{1,\dots,v\}$. A synthetic language is specified by the joint distribution
\begin{align}
    P(\bm{x})
    =
    \mathbb{P}(X_1=x_1,\ldots,X_d=x_d) \ ,
\end{align}
where $X_i$ denotes the random variable at position $i$. In what follows, $P$ varies across different instances of the hierarchical language model. To measure long-range correlations, we use the two-point mutual information. In our hierarchical model, this quantity depends on absolute position, not only on relative distance. We focus on the mutual information between the last token and the token $r$ positions before it,
\begin{align}
    I(r)
    =
    \sum_{a,b\in\mathcal{V}}
    \mathbb{P}(X_{d-r}=a, X_d=b)
    \log_2
    \left[
        \frac{
            \mathbb{P}(X_{d-r}=a, X_d=b)
        }{
            \mathbb{P}(X_{d-r}=a)\mathbb{P}(X_d=b)
        }
    \right]\,,
    \qquad 1\le r\le d-1 \ ,
    \label{eq:mi_definition}
\end{align}
because it provides a clean probe of all hierarchical scales. The logarithm to base 2 measures mutual information in bits. The observable $I(r)$ is our primary statistic for diagnosing whether temperature-rescaled sampling preserves, weakens, or qualitatively changes long-range correlations.

\paragraph{Hierarchical language model.}

We generate the synthetic languages studied in this work with a finite-depth probabilistic context-free grammar (PCFG) on a fixed $s$-ary tree of depth $L$. In PCFG terminology, the observed tokens are terminal symbols, and the internal nodes of the tree are nonterminal symbols. For simplicity, we use the same finite vocabulary $\mathcal{V}=\{1,\dots,v\}$ at every level and track the level by a superscript. Thus $x_i^{(0)}\equiv x_i$ denotes a leaf token, and an observed sequence is $\bm{x} = (x_1,\dots,x_d)$ with $d=s^L$. At height $\ell$, there are $s^{L-\ell}$ nodes $x_i^{(\ell)}$. The root is the single symbol $x_1^{(L)}$, sampled uniformly from $\mathcal{V}$, so each $x_1^{(L)}\in\mathcal{V}$ occurs with probability $1/v$. One may think of the higher-level symbols as higher-level features of a text: a broad latent topic at the root, intermediate subtopics, and emitted tokens at the leaves.

Each internal symbol produces $s$ children one level below. 
The transition from level $\ell$ to level $\ell-1$ is specified by an $(s+1)$-dimensional transition tensor $\bm{\mathsf{M}}^{(\ell)} \in \mathbb{R}^{v\times v\times\cdots\times v}$, where ${\sf M}^{(\ell)}(x,\bm{y}) = {\sf M}^{(\ell)}(x,y_1,\dots,y_s)$ is the probability that a parent symbol $x\in\mathcal{V}$ at level $\ell$ generates the ordered tuple of children $\bm{y}=(y_1,\dots,y_s)\in\mathcal{V}^s$ at level $\ell-1$. For every level and parent symbol, ${\sf M}^{(\ell)}(x,\bm{y}) \in [0,1]$ and $\sum_{\bm{y}\in\mathcal{V}^s} {\sf M}^{(\ell)}(x,\bm{y}) = 1$. Starting from the root, we apply these production rules independently at every internal node for $L$ generations, obtaining the terminal sequence $\bm{x}^{(0)}\equiv \bm{x}$ of length $d=s^L$. This generative process is depicted in \cref{fig:concept}. The transition tensors are drawn from different ensembles of rules.

With this notation, the probability of a complete derivation tree is given by
\begin{align}
    \mathbb{P}\!\left(
        \{\bm x^{(\ell)}\}_{\ell=0}^L
    \right)
    =
    \frac{1}{v}
    \prod_{\ell=1}^{L}
    \prod_{j=1}^{s^{L-\ell}}
    {\sf M}^{(\ell)}
    \left(
        x_j^{(\ell)},
        x_{s(j-1)+1}^{(\ell-1)},
        \dots,
        x_{sj}^{(\ell-1)}
    \right).
\end{align}
The induced language distribution $P(\bm{x})$ is obtained by summing this expression over all internal variables consistent with the leaves $\bm{x}$. When the transition rules are unambiguous, this sum has at most one nonzero contribution. Specifically, for each level $\ell$ and parent symbol $x$, let $\mathcal{T}_x^{(\ell)}\subset\mathcal{V}^s$ denote the set of child tuples that can be generated by $x$. Then, for unambiguous rules,
\begin{align}
    {\sf M}^{(\ell)}(x,\bm{y})
    =
    \begin{cases}
        >0, & \bm{y}\in\mathcal{T}_x^{(\ell)},\\
        0,  & \text{otherwise},
    \end{cases}
\end{align}
with $|\mathcal{T}_x^{(\ell)}|=m \leq v^{s-1}$ and $\mathcal{T}_x^{(\ell)}\cap \mathcal{T}_{x'}^{(\ell)} = \emptyset$ for  $x\neq x'$. 
Equivalently, if
${\sf M}^{(\ell)}(x,\bm{y})>0$, then ${\sf M}^{(\ell)}(x',\bm{y})=0$ for all $x'\neq x$. Thus a tuple of children has at most one possible parent. As a result, for every sequence $\bm{x}$ in the support of the language, the entire latent tree $\big(\bm{x}^{(1)},\bm{x}^{(2)},\dots,x_1^{(L)}\big)$ can be reconstructed deterministically by moving upward from the leaves. While we mainly consider unambiguous ensembles, we also explore ambiguous rules.

The parameter $m$ controls the support of the language. When $m=v^{s-1}$, the sets $\{\mathcal{T}_x^{(\ell)}\}_{x\in\mathcal{V}}$ form a partition of all $v^s$ possible child tuples, so every terminal string has nonzero probability. In this full-support case, correlations arise from heterogeneous rule probabilities: all strings are allowed, but they are not equally likely. In contrast, when $m<v^{s-1}$, some $s$-tuples are forbidden at a given level, and hence some terminal strings have exactly zero probability. In that sparse-support case, even if all allowed rules for a given parent are assigned equal probability, correlations between distant leaves can remain finite because the support itself excludes many strings. In both cases, the fixed hierarchical geometry is the source of long-range structure. Correlations are induced by common ancestors in the tree, and because tree distance grows logarithmically with token separation, the resulting two-point mutual information naturally exhibits power-law envelopes in the leaf distance $r$, as illustrated schematically in \cref{fig:concept}.

\paragraph{Exact next-token inference.}
A central advantage of this hierarchical construction is that the language distribution is known explicitly. For any prefix $(x_1,\dots,x_{i-1})$, we can compute the exact next-token conditional $P(x_i \mid x_1,\dots,x_{i-1})$ without training a model. Because the underlying generative graph is a fixed tree, these conditionals can be computed efficiently by belief propagation~\cite{pearl1982belief,mezard2009information}. For fixed vocabulary size and branching factor, computing the next-token conditional scales linearly in the number of nodes of the derivation tree. This exact inference makes the model a controlled testbed, as it uses the true conditionals of a synthetic language, rather than the output of an imperfectly optimized neural network. At the same time, the setup remains relevant for language-modeling questions, since recent work has shown that transformers trained on related hierarchical languages can learn the exact inference algorithm~\cite{garnier2024hierarchy}.

\paragraph{Temperature-rescaled autoregressive sampling.}
We use these exact conditionals to define temperature-rescaled probabilities. In a large language model, logits $z_a$ over tokens $a\in\mathcal{V}$ are typically converted into sampling probabilities through a softmax with temperature $T>0$, with the probability for token $a$ given by $p_a(T) = \exp(z_a/T)/Z(T)$ where $Z(T) \coloneqq \sum_{b\in\mathcal{V}}\exp(z_b/T)$. Lower temperatures sharpen the distribution, while higher temperatures flatten it. Since in our setting the exact next-token probabilities $P(x_i\mid x_1,\dots,x_{i-1})$ are available by belief propagation, we apply the same transformation directly to the true conditional distribution:
\begin{align}
    P^{(T)}(x_i\mid x_1,\dots,x_{i-1})
    \coloneqq
    \frac{
        P(x_i\mid x_1,\dots,x_{i-1})^{1/T}
    }{
        \sum_{a\in\mathcal{V}}
        P(a\mid x_1,\dots,x_{i-1})^{1/T}
    } \ .
    \label{eq:temperature_conditional}
\end{align}
Thus $T=1$ recovers the original hierarchical language, $T<1$ makes the next-token distribution more peaked, and $T>1$ makes it more uniform. For finite $T$, tokens with zero conditional probability remain forbidden. 
The temperature-rescaled language distribution is then defined autoregressively by multiplying these altered conditionals, $P^{(T)}(\bm{x}) \coloneqq \prod_{i=1}^{d} P^{(T)}(x_i\mid x_1,\dots,x_{i-1})$. 
This construction is the controlled analogue of a language model that has perfectly learned the original language distribution and is then sampled with a decoding temperature different from one. In general, $P^{(T)}(\bm x)$ is not equal to the globally tempered distribution $\propto P(\bm x)^{1/T}$, nor need it correspond to another PCFG obtained simply by altering the transition tensors $\bm{\mathsf{M}}^{(\ell)}$. To study its long-range correlations, we sample complete sequences autoregressively using \cref{eq:temperature_conditional} and estimate the mutual information directly from the resulting samples via the empirical probabilities and \cref{eq:mi_definition}. Due to finite sample size, there is a resolution floor (see \cref{app:additional_results}). Since one next-token inference scales linearly with tree size and we perform it at each of the $d$ positions, the computation time for a single sampled sequence scales quadratically in the sequence length $d$. Except for the exact-inference setup and the later Ising-model interpretation, the core temperature results in this paper are numerical and are obtained from sampled sequences.

\section{Related work}
\label{sec:related_work}

\subsection{Influence of sampling temperature on model outputs}

Sampling temperature is a standard control parameter in autoregressive language models, but its effect on generated text is still not fully understood. Early work on LSTM-generated text showed that temperature can strongly affect word frequencies, entropy measures, and long-range correlations, with an empirically optimal value at which generated text most closely resembles natural language~\cite{lippi2019lstm}. More recent work has extended this perspective to large language models, studying temperature-dependent changes in output distributions and text statistics~\cite{arnold2024phase,mikhaylovskiy2025zipf,kempton2025distortion}, as well as downstream performance and reasoning behavior under temperature shifts~\cite{renze2024temperature,xu2022systematic,zhu2024hot,pursnani2023performance}. These studies emphasize that temperature can induce substantial global changes in the generated sequence distribution.

Recent work has also reported evidence for genuine phase transitions. \citet{bahamondes2023phase} studied GPT-2~\cite{radford2019language} outputs as a function of temperature, using token correlations and statistics of embedding-vector sequences, and reported an onset of deterministic behavior at a relatively low critical temperature of order $T\simeq 0.1$. \citet{nakaishi2024critical} analyzed texts generated by GPT-2 and Pythia~\cite{biderman2023pythia} through statistics of part-of-speech tag sequences of length $512$, reporting signatures of criticality that separate a low-temperature repetitive regime from a high-temperature incoherent regime. Their diagnostics include power-law correlation decay and slow convergence near the transition.

Our work is complementary to this empirical literature. Rather than asking whether trained neural language models exhibit critical behavior, we ask what temperature-rescaled decoding does in a controlled hierarchical language. This removes training error and architectural details while retaining the effect of a local softmax temperature.

\subsection{Related models}

Our model is a finite-depth probabilistic context-free grammar (PCFG), in the tradition of grammar-based models of language~\cite{chomsky1956grammar}. 
For tree-structured generative grammars of this kind, \citet{lin2017critical} showed analytically that the mutual information between two symbols decays as a power law in their separation, through the relation between tree and sequence distances.
We take this mechanism as our starting point and ask how it responds when the exact next-token conditionals of a fixed grammar are rescaled by a decoding temperature. Our model
is also related to the random language model introduced by DeGiuli~\cite{degiuli2019rlm,degiuli2019emergence}. In that model, nonterminals produce pairs of nonterminals with random log-probabilities and can also terminate by producing terminal symbols. The model generates long-range structure, but it differs from ours because the parse tree is not fixed and the hidden nonterminals are not deterministically reconstructable from the observed terminal sequence.

The closest models are the random hierarchy model of Cagnetta~et~al.~\cite{cagnetta2024rhm,cagnetta2024correlations} and the hierarchical model of \citet{garnier2024hierarchy}. These works use fixed hierarchical trees with constant branching ratio and deterministic reconstruction of latent variables from the leaves. They were proposed as controlled models of compositional structure in language and images. In particular, \citet{cagnetta2024correlations} relate long-range token-token correlations in such PCFGs to representation learning and neural scaling behavior. Relatedly, \citet{garnier2024hierarchy} show that transformers trained on their hierarchical language can reconstruct correlations across successive length scales and approximate the exact inference algorithm.

Our setup uses these hierarchical languages as a controlled baseline for studying decoding temperature. Relative to \citet{garnier2024hierarchy}, we retain heterogeneous transition probabilities as the main case, but also consider larger branching ratios and layer-dependent transition tensors. Relative to Cagnetta~et~al.~\cite{cagnetta2024rhm,cagnetta2024correlations}, we compare sparse-support constraints, softened constraints, and full-support heterogeneous rules.

\section{Temperature dependence of the two-point mutual information}
\label{sec:numerics}

\subsection{Full-support rule ensemble with heterogeneous probabilities}

To specify a particular language within the hierarchical model, one must choose the transition probabilities. We first consider a full-support ensemble with heterogeneous probabilities, which is the main setting of the paper. Here $m=v^{s-1}$, so the local rule supports $\{\mathcal{T}^{(\ell)}_x\}_{x\in\mathcal{V}}$ partition the space of child tuples and every terminal sequence has nonzero probability. Correlations are therefore not caused by hard exclusions of complete strings, but by heterogeneous production probabilities. For each allowed rule $\bm y\in\mathcal{T}_x$ we draw an independent standard Gaussian variable $\xi_{x,\bm y}\sim\mathcal{N}(0,1)$ and set
\begin{align}
    {\sf M}(x,\bm y)
    =
    \frac{\exp(\sigma \xi_{x,\bm y})}
    {\sum_{\bm y'\in\mathcal{T}_x}\exp(\sigma \xi_{x,\bm y'})} \ ,
    \qquad
    \bm y\in\mathcal{T}_x \ ,
    \label{eq:lognormal_rules}
\end{align}
with ${\sf M}(x,\bm y)=0$ for $\bm y\notin\mathcal{T}_x$. The transition probabilities are drawn once and then define a fixed language for a given sampling process.  The parameter $\sigma$ controls the heterogeneity of the grammar. We use binary branching, $s=2$, and the same transition tensor for every level, $\bm{\mathsf M}^{(\ell)} \equiv \bm{\mathsf M}$, matching the setting used by \citet{garnier2024hierarchy}.

\begin{figure}[t]
  \centering
  \includegraphics{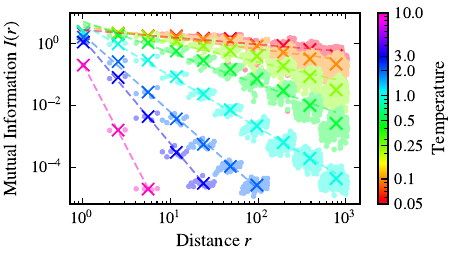}
  \includegraphics{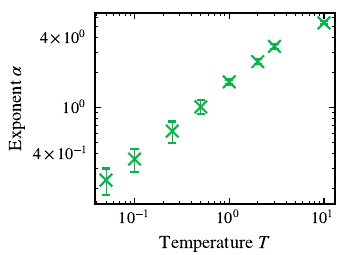}
  \caption{
  Temperature dependence of the two-point mutual information in a grammar drawn from the full-support rule ensemble with heterogeneous probabilities, with $v=16$, $L=10$, and $s=2$, corresponding to sequence length $d=1024$. More details are given in \cref{app:additional_results}.
  Left: Empirical mutual information $I(r)$ as a function of token distance $r$ for several decoding temperatures.
  Points show estimates at individual distances, crosses indicate averages over distances that share the same lowest common ancestor with the last token, and dashed lines show power-law fits.
  Right: Fitted effective exponent $\alpha$ in $I(r)\sim r^{-\alpha}$ as a function of temperature.
  Temperature changes both the magnitude and the exponent of the correlations, but over the tested range the decay remains well described by a power law rather than becoming either a distance-independent plateau or a short-range exponential decay. Error bars indicate $2\sigma$ errors of the fit.}
\label{fig:lognormal_temperature}
\end{figure}

The resulting two-point mutual information curves for a single drawn transition tensor are shown in \cref{fig:lognormal_temperature}, for $L=10$, $v=16$, and $\sigma=5.0$. At $T=1$, sampling reproduces the original hierarchical language. The observed $I(r)$ has the expected staircase structure inherited from the hierarchy: distances $r$ that correspond to the same lowest common ancestor with the last token form a level with comparable correlations, while larger separations probe the next hierarchical scale. We therefore fit the envelope formed by these level-averaged correlations rather than the microscopic fluctuations of the raw curve. Across temperatures, this envelope is well described by an effective power law,
\begin{align}
    I(r) \propto r^{-\alpha(T)} \ ,
\end{align}
over the accessible range of distances. Note, however, that especially at high temperatures the mutual information quickly drops below the finite sample size noise floor,   leaving as few as  three levels for a fit.

Changing the temperature substantially alters the correlations. At low temperature, the next-token conditionals are sharper, the mutual information remains larger at long distances, and the fitted decay is relatively shallow. Increasing temperature flattens the local conditionals, suppresses the overall magnitude of $I(r)$, and increases the fitted exponent. However, this suppression is gradual, and the decay remains well described by a power law. The curves do not exhibit a plateau, as in an ordered phase, nor do they show a crossover to purely short-range exponential decay over the studied system sizes. As $T\to\infty$, $P^{(T)}$ approaches the uniform distribution over sequences and the mutual information must vanish at all distances. However, we find no evidence that, at finite temperatures, the decay takes a form other than a power law.

The exponent plot in \cref{fig:lognormal_temperature} summarizes the same conclusion. The effective exponent $\alpha(T)$ varies smoothly with temperature, increasing continuously from low to high temperatures. This is consistent with a broad crossover in the strength of hierarchical correlations, but it does not suggest a uniquely tuned critical temperature. Note that the displayed $2\sigma$ errors of the fit may underestimate the true uncertainty, because the residuals are not independent across distances. As a further diagnostic, \cref{app:powerlaw_vs_exponential} compares the power-law fit with an exponential alternative on the same level-averaged points and finds that the power-law fit has lower error at all tested temperatures. Somewhat counterintuitively, this preference is strongest at high temperatures rather than low temperatures. Thus, in the main full-support rule ensemble, temperature changes the quantitative decay of the two-point mutual information while preserving an approximate power-law form over a broad range.

\subsection{Sparse support and softened constraints}

As a control, we also study sparse transition rules with equal probabilities, using different transition tensors at different levels to test robustness and facilitate comparison with the random hierarchy model~\cite{cagnetta2024rhm}. Here $m<v^{s-1}$, so each transition can generate only a strict subset of all child tuples. In the strictly sparse version, all allowed rules have probability $1/m$ and all other rules have probability zero, as in the random hierarchy model of \citet{cagnetta2024rhm}. We also consider a softened version in which the grammar allows errors relative to this strict support. Here, with total probability $1-\varepsilon$ the parent emits one of its $m$ allowed tuples uniformly, while with total probability $\varepsilon$ it emits one of the remaining forbidden tuples uniformly. Thus $\varepsilon=0$ gives a language with hard grammatical constraints, whereas $\varepsilon>0$ assigns positive probability to every local production. For $\varepsilon>0$, the latent tree is no longer deterministically reconstructable from the leaves, but exact next-token inference by belief propagation remains available.

\begin{figure}[t]
  \centering
  \includegraphics{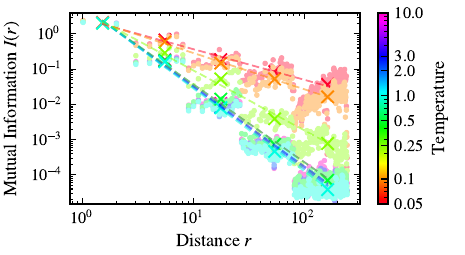}
  \includegraphics{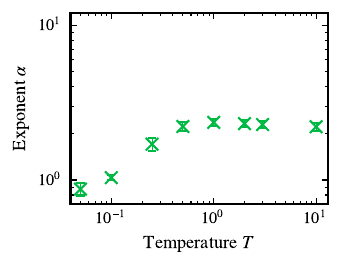}
  \includegraphics{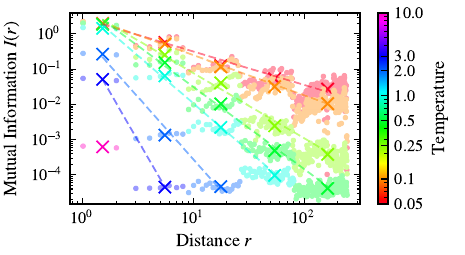}
  \includegraphics{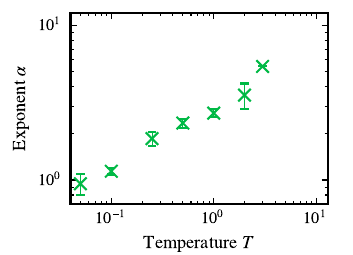}
    \caption{
  Temperature dependence of the two-point mutual information in grammars drawn from sparse rule ensembles with branching factor $s=3$, $L=5$, $v=16$, and $m=4$ allowed transition rules per parent.
  Top row: Strictly sparse grammar with $\varepsilon=0$.
  Bottom row: Softened sparse grammar with $\varepsilon=0.1$.
  In the strict case, forbidden transitions remain forbidden at all finite temperatures and therefore continue to sustain correlations even at high temperature.
  In the softened case, high temperature weakens the preferred sparse structure, but the observed decay remains well described by a power law over the sampled range.
  Additional results for sparse transition rules, including results for $s=2$ and $m=3$, are shown in \cref{app:additional_results}.}
  \label{fig:sparse_temperature}
\end{figure}

The strictly sparse case in the top row of \cref{fig:sparse_temperature} is an informative limiting case. While decreasing the temperature again increases the mutual information, increasing the temperature cannot remove hard grammatical constraints, because a conditional probability of zero remains zero under temperature rescaling. The sampling process is therefore restricted to the same support even at high temperature, and the mutual information remains controlled by the disallowed rules; correspondingly, the power-law exponent saturates.

The softened case in the bottom row of \cref{fig:sparse_temperature} removes this obstruction. Once $\varepsilon>0$, every local transition in the generative process is possible, so high-temperature decoding can gradually flatten the next-token distribution away from the preferred sparse grammar. The behavior is then qualitatively similar to that of the heterogeneous full-support ensemble. Again, increasing temperature suppresses the magnitude of $I(r)$ and steepens the decay, but the decay remains well described by a power law and the change is smooth. We do not observe a sharp feature in $\alpha(T)$, a long-distance plateau, or a clear crossover to exponential decay over the accessible system sizes. Thus the sparse rule ensembles support the same conclusion as the main ensemble: temperature reshapes two-point correlations, but there is no unique critical temperature at which power-law decay first appears.

\section{Power law persistence and the Ising model on trees}
\label{sec:ising_mapping}

We now give a simple statistical-mechanics interpretation of the numerical observations. Consider the special binary vocabulary $v=2$, and identify the two symbols with spins
$\sigma\in\{-1,+1\}$. A minimal transition tensor that favors children aligned with their parent can be written as a Boltzmann weight
\begin{align}
    {\sf M}_{\sigma}(y_1,y_2)
    \propto
    \exp\!\left[
        \frac{J}{2T_{\mathrm{B}}}\sigma(y_1+y_2)
    \right],
    \qquad
    \sigma,y_1,y_2\in\{-1,+1\} \ ,
    \label{eq:ising_rule}
\end{align}
with the normalization performed separately for each parent spin $\sigma$. Here $J>0$ is a ferromagnetic coupling and $T_{\mathrm{B}}$ is the Boltzmann temperature. This rule is exactly the local conditional structure induced by an Ising model on a tree with nearest-neighbor interactions. Low $T_{\mathrm{B}}$ strongly aligns child spins with the parent spin, while high $T_{\mathrm{B}}$ weakens this alignment. Thus, in this simplified limit, the hierarchical language model can be viewed as a spin system on a tree whose leaves are the observed tokens.

This analogy clarifies why lowering the temperature does not necessarily produce an ordered phase in the observed leaf sequence. The underlying geometry is a Cayley tree with open boundary conditions, since the tokens live at the leaves. In particular, it is known that the Ising model on a Cayley tree with open boundaries does not exhibit spontaneous magnetization at any finite temperature~\cite{morita1975cayley}. That is, there is no phase transition to a state with true long-range order at the leaves. The intuitive reason is that, unlike in conventional lattice systems, the boundary of a tree is not negligible: even as the depth grows, a finite fraction of all nodes lies in the outermost generation. Consequently, disorder at the boundary always dominates any ordering tendency in the bulk. In the language-model interpretation, lowering the local noise can strengthen correlations between leaves, but it does not imply a distance-independent mutual-information plateau corresponding to true long-range order.

Instead, in the Ising model defined by \cref{eq:ising_rule}, two leaf positions $i,j$ separated by graph distance $\Delta(i,j)$ on the tree have correlation $\rho(i,j) \coloneqq \langle\sigma_i\sigma_j\rangle = q^{\Delta(i,j)}$, with $q(T_{\mathrm{B}}) = \tanh\!\left[J/(2T_{\mathrm{B}})\right]<1$. Since the one-spin marginals are uniform, two binary spins with correlation $\rho$ have mutual information $I(\rho) \approx \rho^2/(2\ln 2)$ for small $|\rho|$. Thus the pairwise mutual information decays exponentially in tree distance at every finite $T_{\mathrm{B}}$. However, the graph distance between the last token and a previous one grows only logarithmically with sequence distance: for the binary tree, $\Delta(r) \approx 2\ln r/\ln 2$. Therefore, exponential decay in tree distance becomes a power law in sequence distance, $I(r) \sim r^{-\alpha(T_{\mathrm{B}})}$, with $\alpha(T_{\mathrm{B}})=-4\ln q(T_{\mathrm{B}}) / \ln 2$. This is the central geometric reason that power-law correlations persist at any finite temperature and that no exponential decay in sequence distance is observed.

Finally, the temperature in \cref{eq:temperature_conditional} should not be identified literally with the Boltzmann temperature $T_{\mathrm{B}}$ in \cref{eq:ising_rule}. The former is a local transformation applied to exact next-token conditionals, whereas the latter is the global temperature parameter of an equilibrium spin model. The Ising construction is therefore an interpretation of the observed trends, not a formal equivalence for the full autoregressive sampling process. Its quantitative accuracy and range of validity for the full PCFG remain to be established. Relatedly, the absence of a sharp phase transition has also been reported in the so-called random language model~\cite{degiuli2019rlm} for finite vocabularies ($v<\infty$) when varying rule heterogeneity rather than sampling temperature~\cite{nakaishi2022absence}. Taken together, these observations suggest that, in hierarchical languages, power-law two-point correlations arise from the geometry of the tree itself. Temperature mainly renormalizes the strength and effective exponent of these correlations, without an observed transition to a plateau or exponential decay.

\section{Discussion and Limitations}
\label{sec:discussion}

In the hierarchical PCFG studied here, with fixed branching ratio and finite-depth tree geometry, the two-point mutual information remains well described by a power law across the tested range of decoding temperatures. Temperature rescaling changes the sampled distribution in quantitatively important ways: it changes both the overall magnitude of the correlations and the fitted effective exponent. However, within the system sizes and temperatures we study, we do not observe a qualitative transition. The correlations do not become distance-independent, as in an ordered phase, nor do they show a clear crossover to short-range exponential decay. Thus, for this hierarchical language model, power-law behavior is not restricted to a special temperature, and the numerical evidence is more consistent with a broad temperature-dependent crossover than with a sharply tuned phase transition. Consequently, for a fixed-depth tree geometry with fixed branching ratio, power-law correlations alone do not imply that a sampling process is tuned to a unique critical temperature.

Our conclusion is complementary to empirical studies of temperature shifts in trained language models. Such works report substantial temperature-dependent changes in generated text statistics, and in some cases signatures of criticality separating low-temperature repetitive regimes from high-temperature disordered regimes~\cite{bahamondes2023phase,nakaishi2024critical}. Our result does not rule out such phenomena in actual large language models. Instead, it provides a controlled baseline indicating that the hierarchical language studied here does not by itself induce a critical point in the sampling process, and that power-law two-point mutual information can persist across a broad range of decoding temperatures. This is also compatible with the observation of \citet{nakaishi2024critical} that power-law correlations can appear throughout a disordered phase, not only exactly at a transition point, while a crossover to faster decay is reported only at higher temperatures.

A first limitation is that our synthetic language has fixed tree geometry, moderate vocabulary size ($v=16$), and is context-free. This restriction is intentional, as it makes the underlying distribution explicit, allows exact next-token inference via belief propagation, and isolates decoding temperature from training and approximation errors. At the same time, natural language and large language models contain richer forms of dependence than a fixed-tree PCFG. More expressive language models, with varying branching factor or probabilistic production of nonterminals, may therefore exhibit qualitatively different behavior. In particular, recent work on context-sensitive random language models reports a genuine phase transition as a function of a temperature-like parameter of the grammar~\cite{toji2026bkt}. That setting is different from the local softmax-temperature perturbation studied here, but it indicates that moving beyond context-free hierarchical structure may reveal qualitatively different regimes.

A second limitation is that we focus on two-point mutual information. This is the appropriate observable for the question of whether pairwise correlations undergo a phase transition under decoding temperature, but it does not capture all long-context structure. The empirical studies of \cref{sec:related_work} measure correlations through different observables, such as embedding-vector sequences~\cite{bahamondes2023phase} or part-of-speech tags~\cite{nakaishi2024critical}. Our comparison with them is therefore qualitative, at the level of how the correlation decay changes with temperature. Language can contain dependencies distributed across many tokens, and recent work argues that bipartite mutual information can capture additional long-range dependencies beyond two-point statistics~\cite{chen2025mi}. Extending the present controlled temperature study to such observables is a natural next step.

Finally, our use of exact inference rather than a trained transformer is both a limitation and a strength. It is a limitation because it does not model representation learning, optimization dynamics, finite-model approximation, or the architectural details of attention-based networks. It is a strength because it cleanly separates the effect of decoding temperature from those factors. In this sense, the model should be viewed not as a complete model of neural network text generation, but as a baseline. It identifies what can already occur in an explicit hierarchical distribution sampled with temperature-rescaled exact conditionals.

\paragraph{Conclusion.}
We studied temperature-rescaled autoregressive decoding in a tractable hierarchical language model with exact next-token inference. Numerically, we found that two-point mutual information remains well described by a power law across a broad temperature range, while its magnitude and effective exponent vary smoothly with temperature. This controlled baseline provides a starting point for future work on how temperature and language structure interact in richer settings, including multivariate information measures, context-sensitive synthetic languages, and direct comparisons with trained language models.

\emph{Reproducibility. }The code needed to reproduce all results has been made available under \cite{paperrepo}.

\begin{acknowledgments}
M.K. acknowledges financial support by the Federal Ministry of Research, Technology and Space of Germany and by the Sächsische Staatsministerium für Wissenschaft, Kultur und Tourismus in the programme Center of Excellence for AI-research ``Center for Scalable Data Analytics and Artificial Intelligence Dresden/Leipzig'', project identification number: ScaDS.AI.
The work of B.R.~is supported in part by the Simons Foundation Grant No.~12574. 
\end{acknowledgments}

{\small
\bibliographystyle{unsrtnat}
\bibliography{pcfg_powerlaw}
}


\newpage

\appendix

\section{Additional numerical results}
\label{app:additional_results}

\subsection{Sampling and fitting details}
All results in this appendix use the autoregressive sampling protocol described in \cref{sec:setup}. The temperature transformation is also applied to the first-token marginal and to all subsequent exact next-token conditionals as in \cref{eq:temperature_conditional}. Mutual information values are estimates computed from sampled strings. For the power-law fits, distances are grouped according to the level of the lowest common ancestor of positions $d-r$ and $d$, where $d$ is the length of the sequence. The fit is performed on the corresponding level-averaged points. We carry out a linear fit within a log-log plot, fitting
\begin{align}
    \log I(r) = c - \alpha \log r
\end{align}
on the averaged points that remain above the empirical resolution floor. The theoretical finite-sample positive bias of the standard plug-in mutual information estimator under the null hypothesis of independence is approximately $\frac{(v-1)^2}{2N\ln2}$, where $N$ is the number of sampled strings and $v$ is the vocabulary size. With a sample size of $2\times 10^7$, mutual information values are resolved down to approximately $10^{-5}$ bits, in line with the theoretical prediction. At high temperatures and large distances, some averaged points fall below this scale and are not used for fitting. If only two averaged points remain, the exponent is still plotted, but no fit uncertainty is reported. All error bars in exponent plots denote two standard errors from the log-log linear fit.

Regarding the sets of possible child tuples $\{\mathcal T_x^{(\ell)}\}_{x\in\mathcal V}$ for a given grammar, note that they are sampled uniformly subject to the disjointness constraint, i.e., by assigning 
$m$ child tuples without replacement to each parent symbol.

The total computation time for all experiments is about $50{,}000$ CPU-hours.

\subsection{Full-support heterogeneous rules: additional seeds}

\begin{figure}[t]
  \centering
  \includegraphics{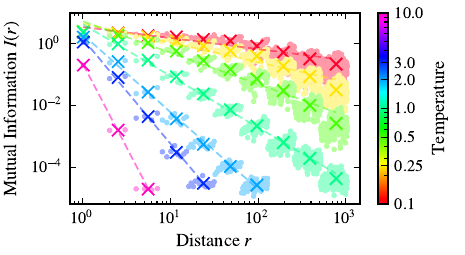}
  \includegraphics{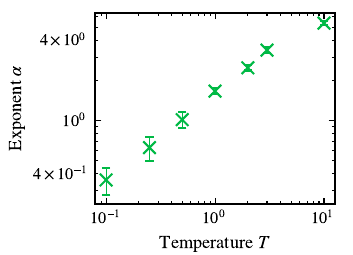}
  \includegraphics{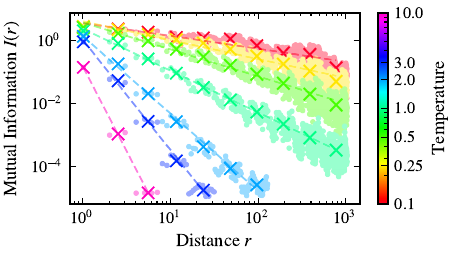}
  \includegraphics{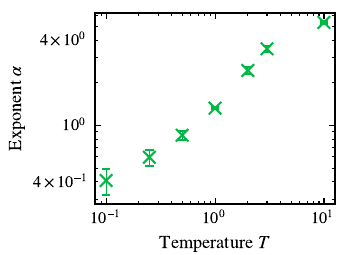}
  \includegraphics{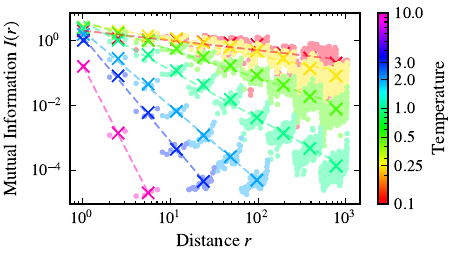}
  \includegraphics{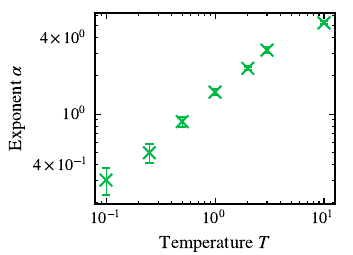}
  \caption{
  Additional temperature sweeps for the full-support heterogeneous rule ensemble with $v=16$, $s=2$, $L=10$, $d=1024$, and $\sigma=5$.
  Rows correspond to three independent rule draws.
  Left column: empirical two-point mutual information $I(r)$ as a function of token distance $r$.
  Points show individual distance estimates, crosses show averages over distances with the same common-ancestor level, and dashed lines show power-law fits to the averaged points.
  Right column: fitted effective exponent $\alpha$ in $I(r)\sim r^{-\alpha}$ as a function of decoding temperature.
  Each curve is estimated from $2.0\times 10^7$ autoregressively sampled strings.
  Error bars denote two standard errors from the fit; missing error bars indicate fits with only two averaged points.
  }
  \label{fig:lognormal_temperature_appendix}
\end{figure}

\cref{fig:lognormal_temperature_appendix} shows additional temperature sweeps for the full-support heterogeneous rule ensemble of \cref{sec:numerics}. We use $v=16$, $s=2$, $L=10$, hence $d=1024$, with the rule heterogeneity of \cref{eq:lognormal_rules} set to $\sigma=5$. The transition tensor is fixed across levels, and the three rows correspond to three independent draws of the transition rules. For each seed and temperature, we sample $N=2.0\times 10^7$ strings autoregressively.

\subsection{Sparse support with binary branching}

\begin{figure}[t]
  \centering
  \includegraphics{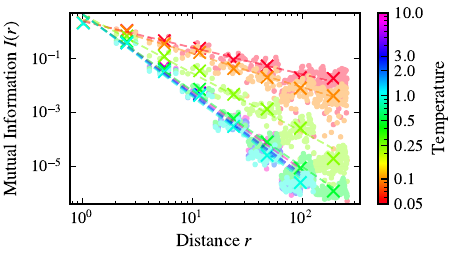}
  \includegraphics{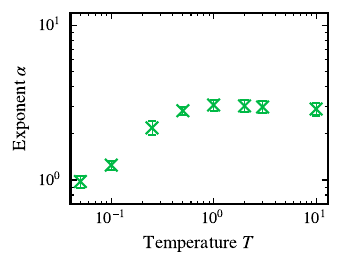}
  \includegraphics{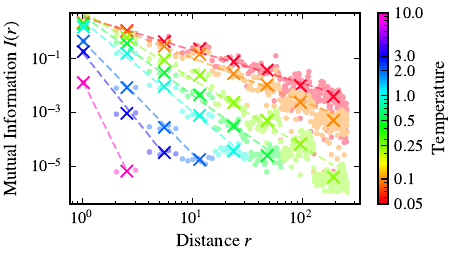}
  \includegraphics{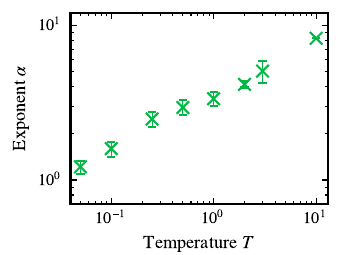}
  \caption{
  Temperature dependence of two-point mutual information in sparse rule ensembles with branching factor $s=2$ and $m=3$ allowed transition rules per parent.
  Top row: strictly sparse grammar with $\varepsilon=0$.
  Bottom row: softened sparse grammar with $\varepsilon=0.1$.
  Left column: empirical mutual information $I(r)$ as a function of token distance $r$.
  Points show individual distance estimates, crosses show common-ancestor-level averages, and dashed lines show power-law fits.
  Right column: fitted effective exponent $\alpha$ in $I(r)\sim r^{-\alpha}$ as a function of decoding temperature.
  Each curve is estimated from $2.0\times 10^8$ autoregressively sampled strings.
  Error bars denote two standard errors from the fit; missing error bars indicate fits with only two averaged points.
  }
  \label{fig:sparse_temperature_s2}
\end{figure}

The sparse-support experiments in the main text use ternary branching, $s=3$, with $v=16$, $L=5$, and $m=4$ allowed transition rules per parent. Those curves are estimated from $1.0\times 10^7$ autoregressively sampled strings for each temperature. As an additional check, we repeat the sparse-support experiment with binary branching, $s=2$, $L=8$, and $m=3$ allowed transition rules per parent. For the binary sparse runs in \cref{fig:sparse_temperature_s2}, each curve is estimated from $2.0\times 10^8$ autoregressively sampled strings, giving a lower resolution floor than in the previous two experiments. Note that, compared to the full-support case above, the transition tensors $\bm{\mathsf{M}}^{(\ell)}$ are different at each level $\ell$ in this setting, to test for any differences and to establish comparability with the random hierarchy model~\cite{cagnetta2024rhm}. 

\subsection{Power-law versus exponential fit diagnostics}
\label{app:powerlaw_vs_exponential}

The main text fits the level-averaged mutual information by a power law. As an additional diagnostic, we compare this fit against an exponential alternative. For the exponential model we fit
\begin{align}
I_{\exp}(r)=A\exp(-r/\xi) \ ,
\end{align}
by a straight line in a semi-logarithmic plot. The power-law model is $I_{\mathrm{pow}}(r)=A r^{-\alpha}$. Both models are fit to the same level-averaged points and over the same resolved distance range used for the exponent estimates in the main text.

To compare the two fits, we evaluate the root mean squared error in logarithmic space,
\begin{align}
\mathrm{RMSE}_{M}
=
\left[
\frac{1}{K}
\sum_{k=1}^{K}
\left(
\log_{10}\bar I(r_k)
-
\log_{10}\widehat I_{M}(r_k)
\right)^2
\right]^{1/2},
\qquad
M\in\{\mathrm{pow},\exp\} \ ,
\end{align}
where $\bar I(r_k)$ denotes the level-averaged empirical mutual information and $\widehat I_M$ denotes the fitted curve. We fit in logarithmic space to capture discrepancies more effectively, including for small mutual information values, and to weight all distances evenly. We report the ratio
\begin{align}
R_{\mathrm{fit}}
=
\frac{\mathrm{RMSE}_{\mathrm{pow}}}{\mathrm{RMSE}_{\exp}} .
\end{align}
Values $R_{\mathrm{fit}}<1$ indicate that the power-law fit gives a smaller logarithmic-scale error than the exponential fit.

\begin{figure}[t]
  \centering
  \includegraphics{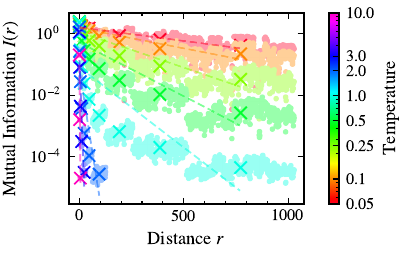}
  \includegraphics{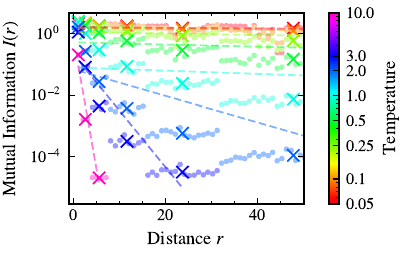}
  \includegraphics{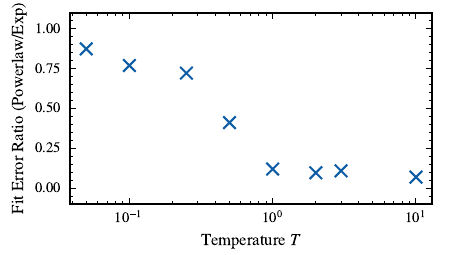}
  \caption{Diagnostic comparison between power-law and exponential fits for the full-support heterogeneous-rule ensemble of \cref{fig:lognormal_temperature}.
Top left: mutual information shown in semi-logarithmic coordinates over the full distance range.
Top right: the same plot zoomed to short and intermediate distances.
In these coordinates, an exponential decay would appear as a straight line, but the level-averaged points, especially at medium and high temperatures, visibly deviate from the fitted exponential lines.
Bottom: ratio of logarithmic-space fit errors, $\mathrm{RMSE}_{\mathrm{pow}}/\mathrm{RMSE}_{\exp}$.
The ratio is below one for all tested temperatures, indicating that the power-law fit is preferred.
  }
  \label{fig:rmse_fit}
\end{figure}

\cref{fig:rmse_fit} shows this comparison for the full-support heterogeneous-rule experiment with the same parameters and seed as in \cref{fig:lognormal_temperature}. In semi-logarithmic coordinates, an exponential decay would appear as a straight line. The empirical level-averaged points do not follow such straight lines, most visibly at intermediate and high temperatures. The RMSE comparison gives the same conclusion quantitatively: for all tested temperatures the power-law fit has lower error than the exponential fit. The preference is strongest for $T\geq 1$, where $R_{\mathrm{fit}}$ is of order $0.1$. At the lowest temperatures the preference is weaker, with ratios closer to $0.8$. This behavior is opposite to the expectation of exponential decay becoming more appropriate in a disordered phase at high temperatures.

\end{document}